%% file: main.tex
\documentclass[10pt,twocolumn,letterpaper]{article}

\usepackage{wacv}              

\input{preamble}
\definecolor{wacvblue}{rgb}{0.21,0.49,0.74}
\usepackage[pagebackref,breaklinks,colorlinks,allcolors=wacvblue]{hyperref}

\title{Generalizing Multimodal Foundational Models in Context-Driven Class Incremental Learning}

\author{Avi Gupta$^1$, Nilotpal Sinha$^2$, Vishnu Raj$^2$, Sambuddha Saha$^2$, Pratik Joshi$^2$,\\ Koteswar Rao Jerripothula$^{3,1}$, Tammam Tillo$^{4,1}$\\
$^1$Indraprastha Institute of Information Technology, Delhi$\quad$ $^2$Dolby Laboratories\\
$^3$Indian Institute of Technology, Kanpur $\quad$ $^4$TYUST, China\\
{\tt\small avig@iiitd.ac.in }
}

\begin{document}
\maketitle
\input{sections/0_Abstract}    
\input{sections/1_Introduction}
\input{sections/2_related}
\input{sections/3_Method}
\input{sections/4_Experiments}
\input{sections/5_Conclusion}
\input{sections/5.5_Acknowledgement}
{
    \small
    \bibliographystyle{ieeenat_fullname}
    \bibliography{main}
}
\clearpage
\input{sections/6_Supplementary}

\end{document}

%% file: preamble.tex
\usepackage{booktabs}       
\usepackage[table]{xcolor}
\usepackage{nicefrac}       
\usepackage{microtype}      
\definecolor{boxcolor}{HTML}{B85450}

\newcommand\crule[3][black]{\textcolor{#1}{\rule{#2}{#3}}}

\newcommand{\cross}{\textcolor{red}{\ding{55}}}

\usepackage{graphicx}
\usepackage{booktabs}
\usepackage{multirow}
\usepackage{enumitem}
\usepackage{comment}
\usepackage{caption} 
\usepackage{arydshln}
\usepackage{tikz}       
\usepackage{adjustbox}  
\usepackage{circledsteps}
\usepackage{amsmath}
\usepackage{wrapfig}
\usepackage{colortbl}
\usepackage{tabularray}
\usepackage{caption}
\usepackage{subcaption}

\usepackage{algorithm}
\usepackage{algpseudocode}
\usepackage{bm}
\usepackage{graphicx}
\usepackage{nicematrix}
\usepackage{bbm}
\usepackage{lipsum}
\usepackage{svg}
\usepackage{wrapfig}
\usepackage{pifont}

\usepackage{url}

\usepackage{pifont} 
\usepackage{xcolor} 

\newcommand{\cmark}{\ding{51}}
\newcommand{\xmark}{\textcolor{red}{\ding{55}}}

\usepackage{amssymb}

%% file: sections/0_Abstract.tex
\begin{abstract}
Class-Incremental Learning (CIL) aims to learn new classes continuously without forgetting the previously learned ones. Recent works in CIL have significantly advanced across various modalities; however, at the intersection of audio-visual modality, this problem remains underexplored. While representations from multimodal foundational models (MFMs) encapsulate rich static priors, our preliminary analysis shows that these representations struggle in incremental settings. Motivated by this, we propose a framework that generalizes frozen MFMs to audio-visual CIL. Specifically, we leverage their dense audio and visual representations and employ an attention-based contextual guidance module, in which the audio representations serve as task-stable contextual guidance for the visual representations, enabling effective class recognition in the incremental setting. To further mitigate catastrophic forgetting and preserve stability across tasks, we employ feature- and logit-level distillation, which are primarily responsible for the reduced forgetting. Extensive evaluations across various audio-visual CIL benchmarks demonstrate that our approach consistently outperforms state-of-the-art methods. Code is available at \href{https://github.com/avigupta2798/MFM-CIL}{https://github.com/avigupta2798/MFM-CIL}.
\end{abstract}

%% file: sections/1_Introduction.tex
\section{Introduction}

\begin{figure*}
\centering
  \includegraphics[width=\textwidth]{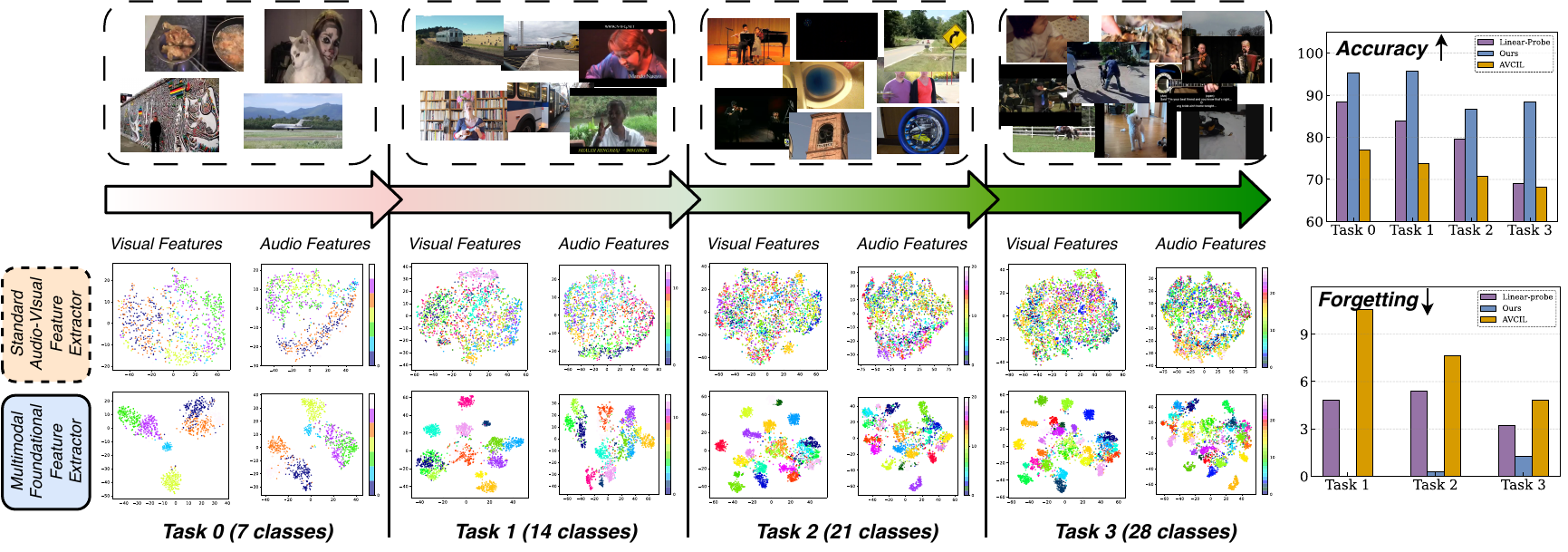}
  \caption{\textit{Comparison of standard audio-visual feature extractor approach with MFMs.} \textbf{Standard approach~\cite{pian2023audio}:} Traditional method processes incoming visual and audio data through stand-alone feature extractors. The t-SNE plots on the AVE dataset from incremental tasks (Task 0 to Task 3) show the overlap and cause clutter, leading to severe catastrophic forgetting, reflected in the decline in testing accuracy [Top-right] and high forgetting [Bottom-right] (\crule[orange!60]{.20cm}{.20cm}). \textbf{MFM-based Approach}: MFMs as feature extractor projects both visual features and audio features into a shared latent space, thereby maintaining clear, distinctly separated class clusters as new tasks arrive and also reflecting in high test accuracy and less forgetting throughout the incremental learning tasks (\crule[blue!55]{.20cm}{.20cm}). \textbf{Linear-Probe} Naive concatenation of MFM's features with no distillation losses and contextual guidance for analyzing the impact of MFMs (\crule[violet!45]{.20cm}{.20cm}).}
  \label{fig:teaser}
\end{figure*}

Class-Incremental Learning (CIL), the task of sequentially learning new classes without succumbing to the catastrophic forgetting of previously learned ones~\cite{zhou2024expandable}, has gained significant interest due to its diverse applications in real-world scenarios, including smart devices, autonomous driving, etc~\cite{yao2025think, yi2025idpa, DBLP:journals/corr/abs-2602-01976, DBLP:journals/corr/abs-2509-20612}. 
Existing literature has primarily focused on strategies such as knowledge distillation~\cite{castro2018end, dhar2019learning, douillard2020podnet}, parameter-efficient adapters~\cite{He_2025_CVPR, wang2025integrating, DBLP:journals/corr/abs-2503-09271, DBLP:journals/corr/abs-2505-11883}, and exemplar-based memory replay~\cite{qi2025class, channappayya2023augmented, chen2023saving}. 
While these approaches have yielded significant advancements in unimodal settings, their efficacy in the context of Multimodal Foundational Models (MFMs) remains largely underexplored. Leveraging cross-modal interactions, particularly the temporal alignment between audio and visual modalities, has been shown to drastically enhance the robustness of latent representations~\cite{DBLP:conf/iccip/LiuL0GL025, DBLP:journals/corr/abs-2505-03186}. However, incrementally leveraging these dense, pre-trained multimodal spaces without disrupting the inter-modal alignment poses a unique and formidable challenge~\cite{DBLP:journals/corr/abs-2512-03125, DBLP:journals/corr/abs-2602-03355}.

Despite the success of MFMs~\cite{radford2021learning, achiam2023gpt, abdin2024phi, kirillov2023segment} in low-level vision tasks~\cite{Yadav_2025_BMVC}, virtual assistants~\cite{kumar2025gpt}, autonomous driving~\cite{guan2024world}, and medical imaging~\cite{azad2023foundational}, MFM-based CIL still faces a \emph{stability-plasticity dilemma}, where the model struggles to balance between learning and integrating new information (plasticity) while retaining previously acquired knowledge (stability)~\cite{DBLP:conf/cvpr/GaoJZZXFDMW25}. Nevertheless, few recent studies~\cite{jha2024clap4clip, cao2024continual, 10656689, wu2025synthetic, DBLP:conf/iccv/ZhengMWQYY23} have explored the use of foundational multimodality, focusing exclusively on Vision-Language foundational models in the class-incremental settings. 
In contrast, other works~\cite{pian2023audio, DBLP:conf/nips/PianNDMGT24, DBLP:journals/corr/abs-2510-17234} incorporate audio-visual inputs through independent feature extractors, and often yield unreliable representations.

A natural question arises: \textit{are MFMs even suitable for incremental learning?} As these models are pretrained on large datasets, they may already possess information relevant to both base and incremental classes. We argue that this apparent tension dissolves once representation priors are distinguished from task knowledge. A frozen encoder supplies generic features, but not a classifier over a specific label set. Catastrophic forgetting arises in the trainable components, whose decision boundaries drift as new classes are introduced. To validate this, we conduct a preliminary study, visualized in Fig.~\textbf{\ref{fig:teaser}}, on the AVE-CI dataset across four incremental tasks. 
We visualize image-specific audio and visual features and observe that, with a standard audio-visual feature extractor (\crule[orange!60]{.20cm}{.20cm}), both modalities overlap and become cluttered as the task progresses. 
This overlap prevents the model from capturing fine-grained semantic cues, as reflected in Fig~\ref{fig:teaser}, through a significant drop in accuracy and high forgetting as it progresses across tasks. 
Conversely, a multimodal foundational feature extractor (\crule[blue!60]{.20cm}{.20cm}) yields significantly clearer class clusters. Yet, stronger features are not sufficient; a sequential linear setting (\crule[violet!60]{.20cm}{.20cm}) on the frozen MFM features, lacking any guidance or distillation, still degrades sharply across tasks (Fig.~\textbf{\ref{fig:teaser}}, $\S$~\ref{sec:linear-probe}). This confirms that strong priors alone do not solve incremental recognition. 
This situates our work within the established paradigm of pre-trained-model-based CIL, where the central question is not whether to leverage frozen foundations, but how to adapt them incrementally without disrupting their cross-modal alignment.
This dual observation motivates us to leverage MFMs in CIL and to actively organize them for incremental learning rather than naively concatenating them.

Hence, we introduce a unified framework that generalizes multimodal foundational models for CIL without requiring full model fine-tuning. We leverage the frozen representations of state-of-the-art MFMs, specifically \textit{SAM-Audio}~\cite{shi2025samaudio}, \textit{ImageBind}~\cite{DBLP:conf/cvpr/GirdharELSAJM23}, and \textit{Qwen2.5-Omni-3B}~\cite{DBLP:journals/corr/abs-2503-20215}, to extract robust, high-dimensional audio and visual features. 
To strategically bridge these modalities, we introduce Attention-based Contextual Guidance that uses extracted audio representations as dynamic context to guide visual features within a transformer-based attention block. 
Unlike prior audio-guided attention, which targets static single-task settings~\cite{Jeong_2025_CVPR}, the incremental setting requires representations to remain stable as new classes arrive; we therefore use audio as task-stable guidance. This results in sample-specific multimodal attended features that capture the fine-grained correlations needed to distinguish novel classes. To mitigate catastrophic forgetting, we further implement the distillation loss (at both the feature and logit levels), constraining the current task's attended features to remain consistent with those of the previous task and preserving the structural integrity of the shared latent space. Finally, a lightweight classifier operates on these stable, integrated representations.

We summarize our key contributions as: {\small \Circled[fill color=black, inner color=white]{\textbf{1}}} We generalize frozen multimodal foundational models (MFMs), specifically \textit{SAM-Audio}, \textit{ImageBind}, and \textit{Qwen2.5-Omni-3B}, to audio-visual class-incremental learning, and we establish that strong MFM priors do not by themselves solve incremental recognition, motivating a targeted adaptation rather than naive feature reuse;
{\small \Circled[fill color=black, inner color=white]{\textbf{2}}} We adapt cross-modal attention, using audio as a task-stable contextual guidance for visual representations in the incremental setting, and show this design generalizes across three distinct MFMs; 
{\small \Circled[fill color=black, inner color=white]{\textbf{3}}} We employ feature- and logit-level distillation to balance the current task's representations with the previous task's state, and further reduce forgetting; 
{\small \Circled[fill color=black, inner color=white]{\textbf{4}}} We conduct extensive experiments on three benchmark datasets, demonstrating consistent improvement over both unimodal and multimodal CIL  baselines.

%% file: sections/2_related.tex
\section{Related Works}
\subsection{Class-Incremental Learning (CIL)}

Class-Incremental Learning (CIL) aims to extend a model's recognition capabilities by continuously learning on new, unknown classes without forgetting the previously learned ones~\cite{jung2020continual, hou2019learning, Kang2022afc, wang2022foster}. 
Existing works primarily address the stability-plasticity dilemma in adapting to new classes while mitigating catastrophic forgetting. 
Parameter regularization-based methods~\cite{ahn2019uncertainty, benjamin2018measuring, Chaudhry_2018_ECCV} penalize weights of critical parameters to indicate their importance. 
Knowledge distillation-based approaches~\cite{castro2018end, dhar2019learning, douillard2020podnet} preserve the trade-offs by maintaining consistency between current and previous model states and minimizing the KL divergence between their respective output distributions. 
Exemplar-based (or Memory Replay) methods~\cite{9710553, chaudhry2018efficient, chaudhry2019tiny, channappayya2023augmented, chen2023saving, DBLP:journals/corr/abs-2509-20612} store a subset of historical exemplar samples to periodically re-train the model, and ensure the decision boundaries remain intact. Lastly, architecture-based methods~\cite{golkar2019continual, hung2019compacting, li2019learn, DBLP:journals/corr/abs-2602-01976} dynamically expand the network's capacity or isolate the task-specific parameters to mitigate inter-task interference.

\subsection{Multimodal Foundational Models}
Recent advancements in multimodal foundation models \cite{radford2021learning, achiam2023gpt, abdin2024phi} have demonstrated remarkable breakthroughs across a spectrum of downstream tasks, including dense prediction \cite{Yadav_2025_BMVC, 10643329, Wu_2024_CVPR}, image recognition \cite{zhang2025multimodal}, and agentic reasoning \cite{sun2025cpathagent}. By aligning disparate modalities into a unified latent space through large-scale pretraining, these architectures excel at capturing cross-modal semantic associations~\cite{wang2024multimodal}. Some notable milestones are SAM-Audio \cite{shi2025samaudio}, ImageBind~\cite{DBLP:conf/cvpr/GirdharELSAJM23}, and Qwen2.5-Omni~\cite{DBLP:journals/corr/abs-2503-20215}. 
These multimodal foundational models share a common latent space across modalities and present a unified representation by synchronizing them. However, despite their zero-shot capabilities, these models are fundamentally constrained by a static learning paradigm. When deployed in non-static environments requiring Class-Incremental Learning (CIL), they remain acutely vulnerable to catastrophic forgetting~\cite{DBLP:conf/iclr/LiuZW025}.

\subsection{Multimodal Foundational Models in CIL}

The emergence of foundational multimodal learning represents a significant paradigm shift in Class-Incremental Learning (CIL)~\cite{huo2025continue, Mo_2023_ICCV, pian2024modality, zhu2024learning, yue2026pal, cui2025few, DBLP:conf/mm/YueZCZLZQ024}. Rather than training task-specific models from scratch, contemporary approaches focus on the sequential adaptation of pre-trained foundations, such as CLIP \cite{DBLP:conf/iclr/GargFPVMTSF24, DBLP:journals/corr/abs-2507-09118, DBLP:journals/corr/abs-2503-08510, DBLP:conf/icml/0001SPFVFTTS25} to non-stationary data distributions \cite{DBLP:conf/iccv/KhattakWNK0K23}. In this setting, the primary objective is to leverage the robust, zero-shot priors of these models while incrementally expanding their semantic understanding without inducing representational drift or degrading the original cross-modal alignment \cite{DBLP:conf/iccv/ZhengMWQYY23, DBLP:conf/nips/GaoZXMW24, DBLP:journals/corr/abs-2507-00469, DBLP:journals/corr/abs-2503-14963, DBLP:conf/icml/LuoZ0Z25}. To navigate the stability-plasticity dilemma, recent works have explored several high-impact strategies, such as memory-based rehearsal for experience replay, Parameter-Efficient Fine-Tuning (PEFT), LoRA \cite{10.1145/3746260.3762126} for low-rank adaptation, and prompt tuning alongside knowledge distillation. Recent analytic and prototype-based approaches built on frozen pretrained models such as RanPAC~\cite{DBLP:conf/nips/McDonnellGPAH23}, GACL~\cite{DBLP:conf/nips/ZhuangC0HTWZC24}, and AnaCP~\cite{DBLP:conf/nips/MomeniXL25} have shown that catastrophic forgetting can be strongly mitigated via closed-form or analytic classifiers. We propose to leverage the foundational multimodal priors and their representations for incremental adaptation, thereby enabling sequential task mastery and retaining prior learning.

%% file: sections/3_Method.tex
\begin{figure*}[ht]
\centering
\includegraphics[width=\textwidth]{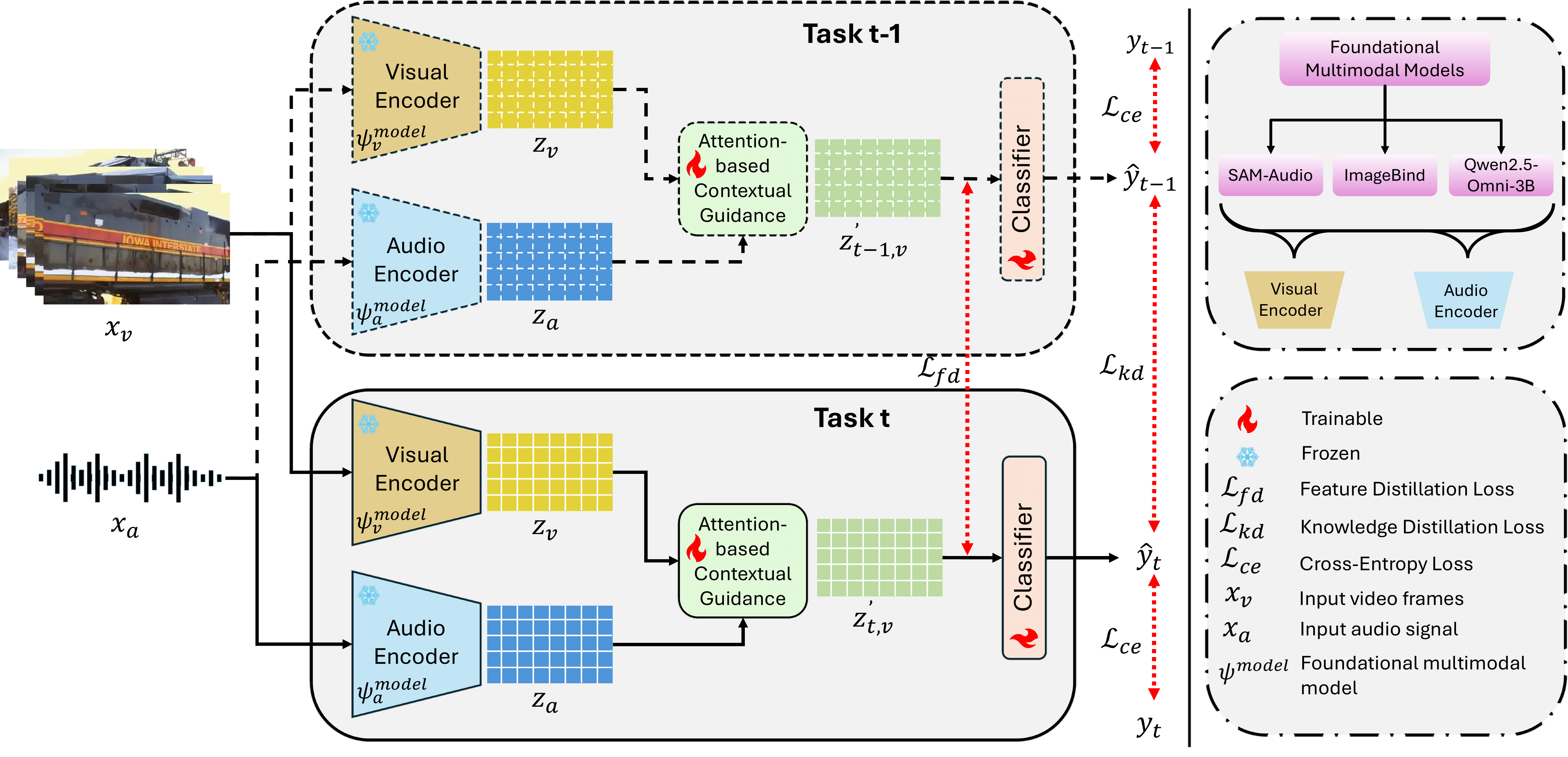}
\caption{\textbf{[Left]} \textit{The overall architecture.} At current task $t$, audio ($z_a$) and visual ($z_v$) representations are derived from audio signal ($x_a$) and video frames ($x_v$) using frozen audio ($\psi^{model}_a$) and visual ($\psi^{model}_v$) encoder, respectively of a Foundational Multimodal model ($model$). The audio representations are utilized as contextual guidance for the corresponding visual representations in a transformer-based attention module. The final video representations ($z'_{t,v}$) are passed through the classifier for precise class prediction ($\hat{y}_t$). The model is trained using $\mathcal{L}_{ce}$ (for class prediction), $\mathcal{L}_{fd}$ (distillation at feature level), and $\mathcal{L}_{kd}$ (distillation at logits level). 
\textbf{[Top-Right]} We leverage three different multimodal foundational models ($model\in\{sam, ibind, qwen\}$) to support our hypothesis.}
\label{fig:arch}
\end{figure*}
\section{Proposed Method}
\subsection{Multimodal Foundational Models (MFMs)}
\label{sec:fmm}
For our work, we leverage three recent foundational multimodal models, namely: \textit{SAM-Audio}, \textit{ImageBind}, and \textit{Qwen2.5-Omni-3B}, to extract audio and visual features.

\textbf{\textit{SAM-Audio}} is a large scale model that extends the ``Segment Anything'' series--originally for images/videos--to the audio domain. The video encoder of the \textit{SAM-Audio} is based on the perception encoder (PE), a vision encoder trained with large-scale contrastive vision-language learning on over 100 million videos. We also leverage the audio encoder of the \textit{SAM-Audio}, which is based on the DAC-like autoencoder structure with a Variational Autoencoder (VAE) bottleneck layer. The audio is encoded into a sequence at 25 Hz.

\textbf{\textit{ImageBind}} is another pioneering multimodal foundational model that unifies six different sensory modalities into a single shared embedding space. In \textit{ImageBind}, both the visual and audio encoders utilize Vision Transformer (ViT) architectures. The model aligns these by treating audio as a 2D image and mapping both into a shared embedding space anchored by visual data. The visual encoder is based on large ViT-H with approximately 630 million parameters. The audio input is first converted into a 2D mel-spectrogram, which is then passed through the ViT-B-based audio encoder. It uses a patch size of 16 and a stride of 10, treating the spectrogram exactly like a 2D image signal.

\textbf{\textit{Qwen2.5-Omni-3B}} is an end-to-end multimodal foundation model. The visual encoder handles both static images and video frames and is based on a Vision Transformer (ViT) encoder with approximately 675 million parameters. The audio encoder is derived from Whisper-large-v3, a state-of-the-art model for speech and sound recognition. Raw audio is resampled to 16 kHz and converted into 128-channel mel-spectrograms using a 25ms window and a 10-ms hop.

For all three models, we only leverage their audio and visual encoders to extract the corresponding features, keeping the weights of all three encoders frozen. 

\subsection{Problem Formulation and Overview}

The incremental learning process is structured as a sequence of tasks $t\in \{0,1,..., T\}$. Each task $t$ introduces a new set of classes $C_t$ that are semantically distinct. The training data available for task $t$ is the dataset $\mathcal{D}_t$. Following~\cite{pian2023audio}, we adopt the updated training dataset for the task $t$, \textit{i.e.} $\mathcal{D}'_t = \mathcal{D}_t\cup \mathcal{M}_{t-1}$, where $\mathcal{M}_{t-1}$ are the exemplar samples, comprising samples from the classes learned so far $C_{t-1}$. These exemplar samples are randomly selected from $C_{t-1}$. Each sample from $\mathcal{D}'_t$ consists of a video sample $x$ and its corresponding ground-truth label $y$. Let $f_t$ denote the training pipeline for the task $t$. At the base task $t=0$, we consider the video sample $(x,y)\in\mathcal{D}_0$ and extract the corresponding video frames $x_v$ and audio signal $x_a$. The $x_v$ and $x_a$ are passed through the visual $\psi_v$ and audio $\psi_a$ encoder to get the corresponding visual $z_v$ and audio $z_a$ encoded features, respectively. These encoded features are fused using attention-based contextual guidance, where the audio-encoded features $z_a$ are used as contextual guidance for visual-encoded features $z_v$ to finally get the contextually-guided features  $z^{'}_v$. These features are finally passed through the classifier to get the predicted label $\hat{y}$. At the incremental task $t>0$, we utilize $\mathcal{D}'_t$, comprising of both current task's samples $\mathcal{D}_t$ and exemplar samples $\mathcal{M}_{t-1}$. For the incremental task, we adopt the same training strategy used in the base task.

In summary, our proposed approach is divided into three major components: \textbf{\textit{(a) Foundational Multimodal Feature Extraction}}, that extracts the audio and visual features from MFM's encoders; \textbf{\textit{(b) Attention-based Contextual Guidance}}, that uses the extracted audio features as contextual guidance for the corresponding visual features in transformer-based attention and provides sample-specific multimodal attended features; and \textbf{\textit{(c) Classifier}}, that predicts the logits from the multimodal attended features. The proposed architecture is visualized in Fig.~\textbf{\ref{fig:arch}}. 

\subsection{Foundational Multimodal Feature Extraction}

As mentioned before, the proposed approach aims to leverage learned audio and visual representations from various foundational multimodal models in an incremental setting. For our approach, we leverage \textit{SAM-Audio} ($sam$), \textit{ImageBind} ($ibind$), and \textit{Qwen2.5-Omni-3B} ($qwen$) as backbones for both the audio and visual modalities. Specifically, we leverage the audio $\psi_a$ and visual $\psi_v$ encoders of the aforementioned models and pass the audio signal $x_a\in\mathbb{R}^{k}$ and visual frames $x_v\in \mathbb{R}^{T_v\times C\times H\times W}$ as inputs to extract the corresponding audio $z_a$ and visual $z_v$ features:
\begin{equation}
\label{eq:feature_ext}
    z^{model}_v = \psi^{model}_v(x_v); \quad z^{model}_a = \psi^{model}_a(x_a);  
\end{equation}

where $model\in\{sam, ibind, qwen\}$, visual features $z^{model}_v \in \mathbb{R}^{T_v\times d^{model}}$, and audio features $z^{model}_a \in \mathbb{R}^{T^{model}_a\times d^{model}}$. $T_{a}^{model}$ and $T_v$ are the audio chunks and visual frames for different multimodal foundational models (MFMs), respectively. $C$, $H$, and $W$ are the number of channels, height and width of each frame, respectively. $d^{model}$ denotes the hidden dimension for different MFMs. We specify the values of $T_v$, $T_a$, and $d$ in Table~\ref{tab:hyperparameters}. 

The extracted features are passed through the attention-based contextual guidance module to leverage the audio representation as context for effective video recognition.

\subsection{Attention-based Contextual Guidance}
The audio $z^{model}_a$ and visual $z^{model}_v$ features from foundational multimodal model ($model\in\{sam, ibind, qwen\}$) represent the sample-wise information. To fuse these features, we perform a preliminary analysis (as illustrated in Table~\textbf{\ref{tab:linear_probe}}) that naively concatenates the audio and visual representations (linear probe); we observe a substantial performance gap. Hence, we propose an attention-based contextual guidance module. This module leverages effective guidance from audio representations to enhance the target object in visual representations. Prior audio-guided attention~\cite{Jeong_2025_CVPR} addresses static, single-task video-text retrieval, where representations need not persist across tasks; our design prioritizes the stability required as new classes arrive. Specifically, the guided attention module is a single-layer attention block comprising layer normalization, Multi-Head Attention (MHA), Feed-Forward Network (FFN), and Multi-Head Self-Attention (MHSA). Below, we provide a detailed breakdown of the pipeline.

First, we pass the normalized visual representations, along with the audio representations (as contextual guidance), into the Multi-Head Attention (MHA) module. In this module, the normalized visual representations serve as the query (Q), while the normalized audio representations serve as the key (K) and value (V), enabling each visual token to attend to the audio context. This setting allows the audio representations to serve as contextual guidance that enhances the visual features, and the attended output preserves the visual sequence length $T_v$, keeping the residual connection well-defined. Following this, the output is fed into a Feed-Forward Network (FFN) with another residual connection. This fusion process is formulated as:
\begin{equation}
\label{eq:attn_begin}
    \begin{aligned}
        \mathbf{z} &= \text{MHA}(\text{LN}(z^{model}_v), \text{LN}(z^{model}_a)) + z^{model}_v; \\
        \overline{\mathbf{z}} &= \text{FFN}_1(\text{LN}(\mathbf{z})) + \mathbf{z}\\
    \end{aligned}
\end{equation}

where LN denotes the layer normalization. MHA denotes the single-layer attention and is formulated as:
\begin{equation}
    \text{MHA(Q,K,V)} = \text{softmax}\Big(\frac{QK^T}{\sqrt{d^{model}}}\Big)V
\end{equation}

where $Q = z^{model}_vW_Q;\ K = z^{model}_aW_K;\ V = z^{model}_aW_V$, and $W_Q, W_K, W_V\in \mathbb{R}^{d^{model}\times d^{model}}$.

The output representation $\overline{\mathbf{z}}$ is further refined to enhance overall visual representations. Specifically, $\overline{\mathbf{z}}$ is passed through the Multi-Head Self-Attention (MHSA) module with another residual connection followed by FFN and final residual connection to produce the refined visual representation 
$z^{'model}_v$. This process is formulated as:
\begin{equation}
\label{eq:attn_end}
    \begin{aligned}
        \tilde{\mathbf{z}} &= \text{MHSA}(\text{LN}(\overline{\mathbf{z}})) + \overline{\mathbf{z}};\\
        z^{'model}_v &= \text{FFN}_2(\text{LN}(\tilde{\mathbf{z}})) + \tilde{\mathbf{z}}
    \end{aligned}
\end{equation}
where MHSA is the single-layer self-attention operation, similar to MHA as $\text{MHSA(X)} = \text{MHA(X, X, X)}$ 
\subsection{Objective Function}
Our proposed strategy employs \textit{cross-entropy} $\mathcal{L}_{ce}$ for class-label prediction. The loss is formulated as:
\begin{equation}
\label{eq:loss_ce}
\mathcal{L}_{ce} = -\sum_{i=1}^{\mathcal{C}_t}y_i\log(\hat{y}_i)
\end{equation}
where $\hat{y}_i$, and $y_i$ denotes the predicted and ground-truth labels respectively. Additionally, we use the \textit{feature distillation loss} $\mathcal{L}_{fd}$ between the attended visual features of the current and the previous task (on the exemplar samples $\mathcal{M}_{t-1}$) to teach $z^{'model}_{v,t}$ to mimic $z_{v,t-1}^{'model}$. It is formulated as:
\begin{equation}
\mathcal{L}_{fd} = ||z^{'model}_{v,t}-z_{v,t-1}^{'model}||^2
\end{equation}
The last part of our final objective function is the \textit{task-wise knowledge distillation} $\mathcal{L}_{kd}$, where we evaluate the Kullback-Leibler (KL) divergence between the logits of classes in the current ($\hat{y}_t^{C_s}$) and previous ($\hat{y}_{t-1}^{C_s}$) tasks which helps preserve past knowledge and prevent the learned knowledge from being biased by other tasks.
\begin{equation}
\mathcal{L}_{kd} = \sum_{s=1}^t KL(\hat{y}_t^{C_s} || \hat{y}_{t-1}^{C_s})
\end{equation}
To sum it up, our overall \textbf{\textit{objective function}} is formulated as: $\mathcal{L}_{tot} = \mathcal{L}_{ce} +\lambda_{fd}*\mathcal{L}_{fd} + \lambda_{kd}*\mathcal{L}_{kd}$, where
$\lambda_{fd}$ and $\lambda_{kd}$ are the hyperparameters for feature and knowledge distillation losses, respectively. 

%% file: sections/4_Experiments.tex
\section{Experiments}
\subsection{Experimental Details}
\paragraph{Datasets.}To ensure fair comparison, we follow AV-CIL~\cite{pian2023audio} and evaluate our proposed approach on three class-incremental learning (CIL) datasets: AVE-Class-Incremental (AVE-CI) dataset~\cite{DBLP:conf/eccv/TianSLDX18}, which comprises over 4,000 video sequences covering 28 distinct event categories. We divide the 28 classes into four sequential tasks, each containing 7 classes. The Kinetics-Sounds-Class-Incremental (K-S-CI) dataset~\cite{DBLP:journals/corr/KayCSZHVVGBNSZ17} consists of 23,000 samples spanning 30 audio-visual classes. The dataset is divided into 5 incremental tasks, each containing 6 classes. The VGGSound100-Class-Incremental (VS100-CI) dataset~\cite{DBLP:conf/icassp/ChenXVZ20} is a large-scale benchmark comprising 60,000 samples across 100 classes. These classes are distributed across 10 incremental tasks, with 10 classes per task, providing a more rigorous test of long-term catastrophic forgetting.

\paragraph{Evaluation Metrics.}
We employ two standard metrics used in CIL: \textit{Mean Accuracy} and \textit{Average Forgetting}.

\textbf{Mean Accuracy ($\mathcal{A}_{m}$). }This metric provides a global measure of the model's performance across all observed categories. After the model has been trained on $T$ sequential tasks, the Mean Accuracy is defined as the mean of the cumulative accuracies recorded at each incremental task:
\begin{equation}
    \mathcal{A}_{m} = \frac{1}{T}\sum_{t=1}^{T}\mathcal{A}_t
\end{equation}
where $\mathcal{A}_t$ is the test accuracy of all seen classes after completing the training on the current task $t$ (\textit{i.e.}, $\bigcup_{i=1}^t \mathcal{C}_i$).

\textbf{Average Forgetting $\mathcal{F}_{avg}$. }To evaluate the impact of catastrophic forgetting on previously learned knowledge, we employ the Average Forgetting. For a specific task $t$, the forgetting $\mathcal{F}_t^i$ regarding a previous task $i<t$ is defined as the maximum decrease in accuracy observed for that task up to the current state. The Average Forgetting after $T$ tasks is expressed as:
\begin{equation}
\label{eq:forget}
\begin{aligned}
    \mathcal{F}_{avg} &= \frac{1}{T-1}\sum_{t=2}^{T}\mathcal{F}_t \\
    \mathcal{F}_t = \frac{1}{t-1}\sum_{i=1}^{t-1}& \max_{k\in\{i,...,t-1\}}(\mathcal{A}_{k,i}- \mathcal{A}_{t,i})
\end{aligned}
\end{equation}
where $\mathcal{A}_{k,i}$ is the test accuracy of the i-th task after training on the k-th task. 

\paragraph{Implementation Details.}We implemented our framework using the PyTorch framework on a single workstation with a \input{Tables/hyperparameters}NVIDIA A100 GPU. In this work, we leverage three different foundational models \textit{SAM-Audio}~\cite{shi2025samaudio}, \textit{ImageBind}~\cite{DBLP:conf/cvpr/GirdharELSAJM23}, and \textit{Qwen2.5-Omni-3B}~\cite{DBLP:journals/corr/abs-2503-20215} for their pre-trained audio and visual encoders. During training, we freeze the visual and audio encoders of all the foundational models and train the rest of the pipeline. We use the SGD optimizer with a learning rate of 5e-5, weight decay of 1e-4, and a Cosine Annealing scheduler. For all three datasets, we set the maximum number of training epochs in each incremental task to 200. Following AV-CIL~\cite{pian2023audio}, we set exemplar memory size to 340 for AVE-CI, 500 for K-S-CI, and 1500 for VS100-CI. 
$\lambda_{fd}$, and $\lambda_{kd}$ are set to 1.0. The Attention-based Contextual Guidance module uses a single-layer block with 16 attention heads and an FFN of hidden dimension $d^{model}$. The classifier is a single fully connected layer that maps the attended features to the number of seen classes, expanded incrementally as new tasks arrive. 
Additional details are provided in Table\textbf{~\ref{tab:hyperparameters}}.

\subsection{Results and Analysis:}
\begin{figure*}[htb]
\centering
\includegraphics[width=\linewidth]{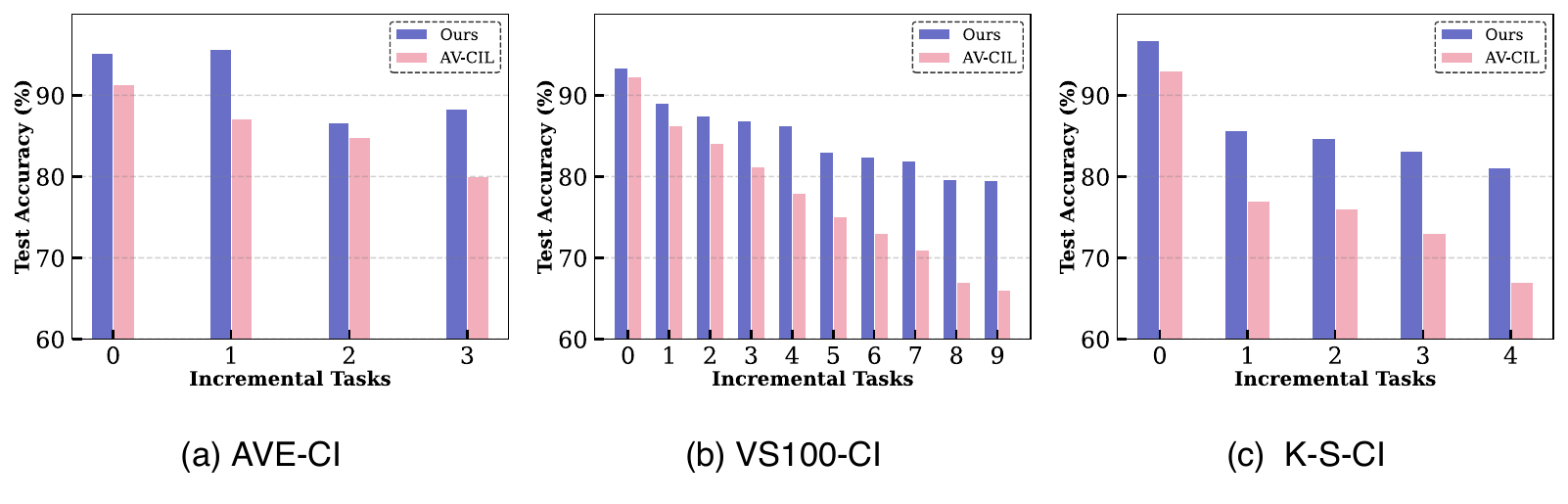}
\caption{\small Comparative analysis of task-wise test accuracy ($\%$) between AV-CIL~\cite{pian2023audio} and our proposed approach (on ImageBind). (a) Individual task performance for the AVE-CI dataset on 4 incremental tasks. (b) Individual task performance for the VS100-CI dataset on 10 incremental tasks. (c) Individual task performance for the K-S-CI dataset on 5 incremental tasks.}
\label{fig:main_analysis}
\end{figure*}
\input{Tables/results}
\input{Tables/fair_results}
\subsubsection{\textbf{Comparison with State-of-the-Art}}

We evaluate our proposed framework along two cues: (a) a backbone-matched evaluation; (b) a generalized evaluation on the three multimodal foundational models. Across both cues, we leverage corresponding extracted features and hyperparameters, and we randomly select exemplar samples for both ours and AV-CIL (using published configuration). 

\textbf{Backbone-matched comparison. }For fair comparison, following AV-CIL~\cite{pian2023audio}, we leverage the AudioMAE~\cite{DBLP:conf/nips/000100BAGMF22} and VideoMAE~\cite{DBLP:conf/nips/TongS0022} encoders in our pipeline. As summarized in Table\textbf{~\ref{tab:results}}, our approach consistently outperforms existing baselines by a substantial margin across all datasets. Specifically, we observe a consistent improvement of \textbf{+0.5\%}, \textbf{+1.57\%}, and \textbf{+0.87\%} on AVE-CI, VS100-CI, and K-S-CI, respectively, from MMAL~\cite{DBLP:conf/mm/YueZCZLZQ024} (under the non-exemplar setting), and against the AV-CIL baseline (for exemplar setting), the margin widens to \textbf{+4.63\%}, \textbf{+6.32\%}, and \textbf{+8.15\%}. This indicates that even without strong pretrained features, our contextual guidance and dual distillation yield a better stability-plasticity trade-off than the previous state of the art.

\textbf{Generalization across MFMs. }To generalize across different multimodal foundational models (MFMs), we comprehensively compare our approach against AV-CIL~\cite{pian2023audio} on three recent MFMs, \textit{SAM-Audio}, \textit{ImageBind}, and \textit{Qwen2.5 Omni}. From Table\textbf{~\ref{tab:fair_results}}, we observe that our approach shows significant improvement across all datasets. The gains are largest on \textit{ImageBind}, where we improve by \textbf{+5.68\%}, \textbf{+7.24\%}, and \textbf{+8.33\%} in accuracy on AVE-CI, VS100-CI, and K-S-CI with similar improvement holding for \textit{SAM-Audio} and \textit{Qwen2.5} as well. Furthermore, our \textit{ImageBind}-based approach achieves a mean accuracy gain of \textbf{$\approx$3\%} over the other two MFMs across the three datasets.
When compared to AV-CIL, our approach improves by \textbf{$\approx$7\%} in mean accuracy on the AVE-CI and K-S-CI datasets. 

\begin{wrapfigure}{l}{0.45\linewidth}
  \centering
  \includegraphics[width=\linewidth]{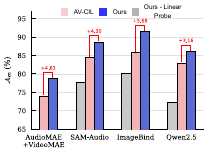}
  \caption{\small Analysis of AV-CIL \& ours on different backbones (AVE-CI).}
  \label{fig:disentangling}
\end{wrapfigure}To further validate the stability of our approach, we provide a task-wise test accuracy analysis in Fig. \textbf{\ref{fig:main_analysis}}, comparing our approach against AV-CIL on \textit{ImageBind} backbone. While AV-CIL suffers from performance decay as the number of tasks increases, our proposed approach maintains nearly consistent performance across tasks.

\textbf{Analysis on Backbone Contributions: }

Fig.~\ref{fig:disentangling} visualizes different backbones and the resulting Ours--AV-CIL margin. We observe that, with the same conventional (Audio/Video MAE) backbone, our method already surpasses AV-CIL by +4.63\% on AVE-CI. This improvement is attributable solely to the architecture, since both methods use the same features, and is observed across other MFM-based backbones as well. Furthermore, the performance improvement is not biased toward our method. AV-CIL's configuration transfers well to MFM features, and in several settings, its improvement exceeds ours (\emph{eg} VS100/Qwen: +7.63 \emph{vs} +3.28; K-S/SAM: +7.15 \emph{vs} +2.80). 

This margin indicates that the improvement is from the proposed contextual guidance and distillation rather than from backbone-specific tuning. 

\subsubsection{Impact of MFMs in Linear Setting}
\label{sec:linear-probe}

In this section, we evaluate the impact of MFMs by training a linear probe on each MFM (naive concatenation). We train a single linear classifier on the frozen audio-visual features, task by task, using only cross-entropy, without contextual guidance or distillation. As shown in Table\textbf{~\ref{tab:linear_probe}}, the linear probe performs poorly across all backbones and datasets, reaching only \textbf{$\approx$73\%} mean accuracy on ImageBind (far below our full method \textbf{$\approx$87\%}) and degrading sharply across tasks (Fig.\textbf{~\ref{fig:teaser}}). This indicates that without proper guidance, the frozen backbone’s representations are susceptible to inter-modality interference. Notably, the probe performs poorly even against AV-CIL under the same conditions, indicating that frozen MFM features are not directly usable for incremental learning without targeted adaptation. This confirms that strong priors alone do not solve the task. 

\input{Tables/linear_probe}

\begin{figure*}[ht]
\centering
\includegraphics[width=\linewidth]{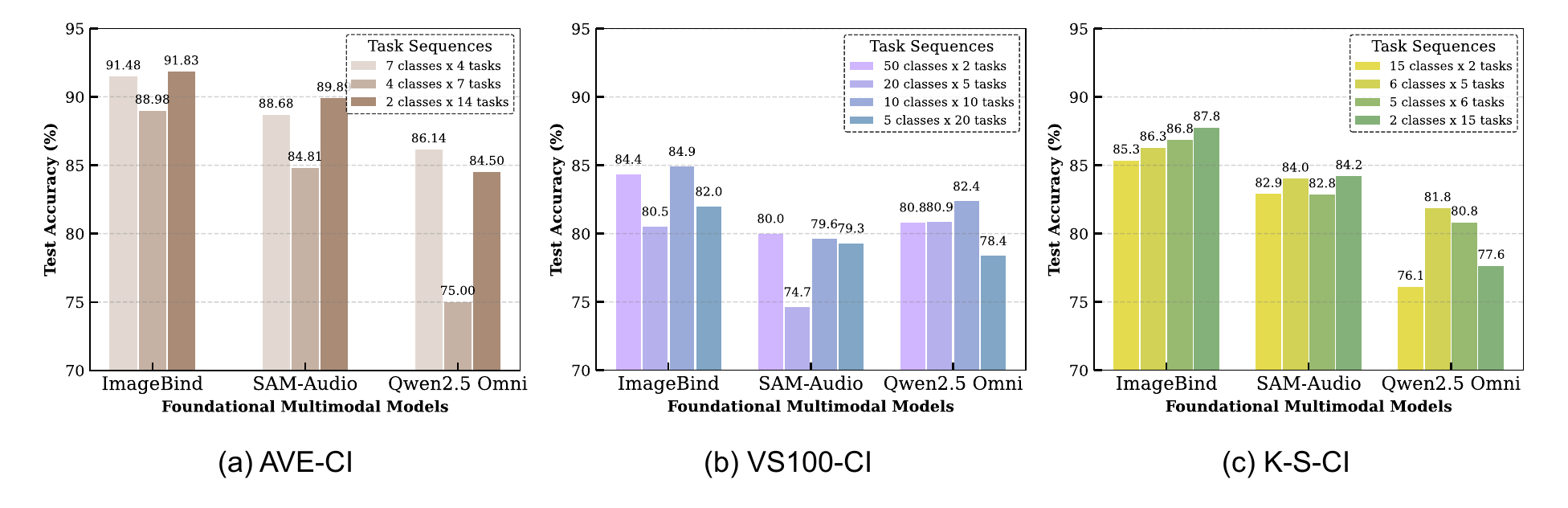}
\caption{Comparative analysis among MFM-based approaches for different task sequences using mean test accuracy ($\%$). (a) Three task sequences for AVE-CI, 4 tasks (7 classes per task), 7 tasks (4 classes per task), and 14 tasks (2 classes per task); (b) Four task sequences for VS100-CI, 2 tasks (50 classes per task), 5 tasks (20 classes per task), 10 tasks (10 classes per task), 20 tasks (5 classes per task); (c) Four task sequences for K-S-CI, 2 tasks (15 classes per task), 5 tasks (6 classes per task), 6 tasks (5 classes per task), 15 tasks (2 classes per task).}
\label{fig:analysis}
\end{figure*}
\subsubsection{\textbf{Robustness to Task Sequences}}
To evaluate the robustness of our proposed approach, we conduct a comprehensive analysis across varying incremental task sequences for all three datasets.
For the AVE-CI dataset, we introduce additional 2-task settings (4 classes per task and 2 classes per task). For K-S-CI, we introduce additional 3-task settings (15 classes per task, 5 classes per task, and 2 classes per task). Finally, for VS100-CI, we introduce three additional task settings (50 classes per task, 20 classes per task, and 5 classes per task). The results are summarized in Fig.\textbf{~\ref{fig:analysis}}.
Our proposed approach demonstrates remarkable stability, maintaining high performance regardless of whether the model encounters large tasks or many small, frequent tasks. This consistent performance across diverse task sequences underscores the architectural robustness of our approach. A detailed task-wise comparative analysis with AV-CIL~\cite{pian2023audio} is presented in the {\color {wacvblue}\textit{\textbf{supplementary}}} file.

\input{Tables/ablation_loss_upd}
\input{Tables/ablation_hyp}
\subsubsection{Effectiveness of Individual Loss Functions}

To analyze the contributions of each component, we conduct an ablation study on the AVE-CI benchmark. We specifically evaluate the impact of Knowledge Distillation ($\mathcal{L}_{kd}$) and Feature Distillation ($\mathcal{L}_{fd}$) relative to the standard Cross-Entropy ($\mathcal{L}_{ce}$) baseline. The results are summarized in Table\textbf{~\ref{tab:ablation_loss_upd}}. Comparing the baseline ($\mathcal{L}_{ce}$) to the inclusion of $\mathcal{L}_{kd}$ shows a consistent improvement in Mean Accuracy (from \textbf{86.55\%} to \textbf{87.60\%} for \textit{ImageBind}), and forgetting (from \textbf{3.18\%} to \textbf{2.80\%}). $\mathcal{L}_{kd}$ serves as a regularizer that maintains the decision boundaries of previously learned classes, hence preventing the classifier from over-fitting to the distribution of the current task. Introducing $\mathcal{L}_{fd}$ provides further improvement than $\mathcal{L}_{kd}$ alone. While $\mathcal{L}_{kd}$ manages label-space consistency, $\mathcal{L}_{fd}$ ensures feature stability (reducing forgetting from \textbf{3.18\%} to \textbf{1.02\%}). The combination results in a jump in accuracy (\textbf{91.48\%}) and a drop in forgetting (\textbf{0.52\%}). We observe a similar trend in other MFMs as well.

\subsubsection{Sensitivity Analysis of Hyperparameters}
To analyze the contributions of our distillation components, we conduct a study over $\lambda_{fd}$ and $\lambda_{kd}$. These values balance the trade-off between preserving feature representations and transferring logit-level knowledge. The results, summarized in Table\textbf{~\ref{tab:ablation_hyp}}, provide insights into the stability-plasticity balance in our framework. We observe that, when $\lambda_{fd}$ and $\lambda_{kd}$ are both set to \textbf{1.0} provides the best overall trade-off, achieving the highest Mean Accuracy in five of the six columns. The only exception is K-S-CI with SAM-Audio, where $\lambda_{fd}/\lambda_{kd}=0.2/1.0$ attains 84.50 \textit{vs} 84.01 at the adopted setting. We retain $1.0/1.0$ as it is optimal for the remaining configurations and best on average. For a given backbone and dataset, the variance between configurations remains small, indicating low sensitivity to the weights.

%% file: Tables/hyperparameters.tex
\setlength{\columnsep}{8pt}
\begin{wraptable}{l}{130pt}
\centering
\vspace{-13pt}
\caption{Architecture details.}
\label{tab:hyperparameters}
\vspace{-8pt}
\resizebox{135pt}{!}{
\centering
\setlength{\tabcolsep}{7pt}
\begin{tabular}{l c c c}
\toprule \midrule
Models (model) &$d^{model}$ & $T_v$ & $T^{model}_a$ \\
\midrule
SAM-Audio &   1024 & 16 & 59 \\
ImageBind & 1024 & 16 & 1 \\
Qwen2.5-Omni-3B &  2048  & 16 & 251 \\
\midrule \bottomrule
\end{tabular}
}
\vspace{-9pt}
\end{wraptable}

%% file: Tables/results.tex
\begin{table}[ht]
\centering
\caption{Comparison of our proposed approach with previous works on different datasets. $\dagger$	represents results that are produced for fair comparison. Best results ($\%$) are marked in \textbf{Bold}.}
\label{tab:results}
\renewcommand{\arraystretch}{1}
\resizebox{\columnwidth}{!}{
\setlength{\tabcolsep}{5pt}
\begin{tabular}{lccccccc}
\toprule
\midrule
\multirow{2}{*}{Methods} & \multirow{2}{*}{Exemplar} & \multicolumn{2}{c}{AVE-CI} & \multicolumn{2}{c}{VS100-CI} & \multicolumn{2}{c}{K-S-CI} \\
\cmidrule(lr){3-4} \cmidrule(lr){5-6} \cmidrule(lr){7-8}
& & $\mathcal{A}_{m}\uparrow$ & $\mathcal{F}_{avg}\downarrow$ & $\mathcal{A}_{m}\uparrow$ & $\mathcal{F}_{avg}\downarrow$ & $\mathcal{A}_{m}\uparrow$ & $\mathcal{F}_{avg}\downarrow$ \\
\midrule
Fine-tuning   & \cross        & 42.40 & 70.99 & 26.21 & 89.37 & 41.18 & 89.62 \\
LwF~\cite{8107520}    & \cross               & 58.07 & 26.90 & 59.34 & 23.01 & 65.54& 16.55 \\
iCaRL-NME~\cite{8100070}     & \checkmark        & 56.15 & 11.71 & 56.19 & 12.80 & 64.51 & 18.70 \\
iCaRL-FC~\cite{8100070}   & \checkmark           & 65.88 & 26.08 & 64.22 & 29.94 & 65.54 & 40.57 \\
SS-IL~\cite{9710553}       & \checkmark          & 61.94 & 22.49 & 69.20 & 9.75 & 69.71 & 10.53 \\
AFC-NME~\cite{Kang2022afc}     & \checkmark          & 68.46 & 14.18 & 61.41 & 23.30 &  69.13 & 27.30 \\
AFC-LSC~\cite{Kang2022afc}     & \checkmark          & 65.21 & 28.11 & 57.76 & 29.64  &  67.02 & 33.56 \\
AV-CIL~\cite{pian2023audio}     & \checkmark           & 74.04 & 7.63  & 72.80 & 5.49 &  73.06 & 6.48  \\ 
MMAL~\cite{DBLP:conf/mm/YueZCZLZQ024}  & \cross  & 76.71 & -  & 76.19 & - & 78.98 & - \\
CIGN$^\dagger$~\cite{Mo_2023_ICCV}  & \cross  & 75.78 & 6.88  & 73.11 & 5.67 & 74.08 & 6.45 \\
MMCL~\cite{10.1145/3746266.3762161}  & \checkmark  & 75.07 & 6.51  & 74.37 & 6.36 & 73.62 & 7.44 \\
PromptCIL~\cite{thakur5929213promptcil}  & \checkmark  & 71.11 & 3.12  & 74.14 & {5.47} & 73.34 & 5.70 \\
CMCDR~\cite{DBLP:journals/corr/abs-2607-27260}  & \cross  & 74.63 & {14.02}  & - & {-} & 75.31 & 9.86 \\
\midrule
\rowcolor{wacvblue!15}
{Ours} & {\cross}   & {77.21} & {3.08}  & {77.76} & {5.12} & {79.85} & {6.02} \\
\rowcolor{wacvblue!15}
Ours & \checkmark   & \textbf{78.67} & \textbf{1.05}  & \textbf{79.12} & \textbf{4.96} & \textbf{81.21} & \textbf{5.26} \\
\midrule
\bottomrule
\end{tabular}
}
\end{table}

%% file: Tables/fair_results.tex
\begin{table}[ht]
\vspace{-5pt}
\centering
\caption{\small Comparison of our proposed approach on different MFMs with AV-CIL~\cite{pian2023audio}. $\dagger$	represents results that are produced for fair settings. Best results ($\%$) are marked in \textbf{Bold}.}
\label{tab:fair_results}
\vspace{-8pt}
\renewcommand{\arraystretch}{1.1}
\resizebox{.85\columnwidth}{!}{
\setlength{\tabcolsep}{3pt}
\begin{tabular}{l cc cc cc}
\toprule
\midrule
\multirow{2}{*}{Methods} & \multicolumn{2}{c}{AVE-CI} & \multicolumn{2}{c}{VS100-CI} & \multicolumn{2}{c}{K-S-CI} \\
\cmidrule(lr){2-3} \cmidrule(lr){4-5} \cmidrule(lr){6-7}
& $\mathcal{A}_{m}\uparrow$ & $\mathcal{F}_{avg}\downarrow$ & $\mathcal{A}_{m}\uparrow$ & $\mathcal{F}_{avg}\downarrow$ & $\mathcal{A}_{m}\uparrow$ & $\mathcal{F}_{avg}\downarrow$ \\
\midrule
AV-CIL$^\dagger$ (SAM-Audio)                & 84.38 & 2.24  & 74.04 & 4.11 &  80.21 & 1.45  \\
\rowcolor{wacvblue!15}
Ours (SAM-Audio)                & \textbf{88.68} & \textbf{0.42}  & \textbf{79.64} & \textbf{0.34} & \textbf{84.01} & \textbf{0.43} \\ \midrule
AV-CIL$^\dagger$ (Qwen2.5)                & 82.98 & 5.91  & 80.43 & 5.40 &  79.32 & 4.28  \\
\rowcolor{wacvblue!15}
Ours (Qwen2.5)                & \textbf{86.14} & \textbf{4.18} & \textbf{82.40} & \textbf{1.20} & \textbf{81.85} & \textbf{3.84} \\ \midrule
AV-CIL$^\dagger$ (ImageBind)                & 85.80 & 3.04  & 77.70 & 7.72 &  77.96 & 7.09  \\ 
\rowcolor{wacvblue!15}
Ours (ImageBind)               & \textbf{91.48} & \textbf{0.52} & \textbf{84.94} & \textbf{0.06} & \textbf{86.29} & \textbf{0.44} \\

\midrule
\bottomrule
\end{tabular}
}
\vspace{-15pt}
\end{table}

%% file: Tables/linear_probe.tex
\begin{table}[ht]
\centering
\caption{\small Comparison of our proposed approach on different MFMs with the naive concatenation (Linear Probe). Best results ($\%$) are in \textbf{Bold}.}
\label{tab:linear_probe}
\renewcommand{\arraystretch}{1.1}
\resizebox{.8\columnwidth}{!}{
\setlength{\tabcolsep}{3pt}
\begin{tabular}{l cc cc cc}
\toprule
\midrule
\multirow{2}{*}{Methods} & \multicolumn{2}{c}{AVE-CI} & \multicolumn{2}{c}{VS100-CI} & \multicolumn{2}{c}{K-S-CI} \\
\cmidrule(lr){2-3} \cmidrule(lr){4-5} \cmidrule(lr){6-7}
& $\mathcal{A}_{m}\uparrow$ & $\mathcal{F}_{avg}\downarrow$ & $\mathcal{A}_{m}\uparrow$ & $\mathcal{F}_{avg}\downarrow$ & $\mathcal{A}_{m}\uparrow$ & $\mathcal{F}_{avg}\downarrow$ \\
\midrule
Linear Probe (SAM-Audio)     & 77.62 & 3.43  & 61.31 & 4.75 &  79.71 & 2.87  \\
\rowcolor{wacvblue!15}
Ours (SAM-Audio)                & \textbf{88.68} & \textbf{0.42}  & \textbf{79.64} & \textbf{0.34} & \textbf{84.01} & \textbf{0.43} \\ \midrule
Linear Probe (Qwen2.5)     & 72.30 & 9.13  & 73.36 & 6.15 &  72.84 & 7.31  \\
\rowcolor{wacvblue!15}
Ours (Qwen2.5)                & \textbf{86.14} & \textbf{4.18} & \textbf{82.40} & \textbf{1.20} & \textbf{81.85} & \textbf{3.84} \\ \midrule
Linear Probe (ImageBind)      & 80.17 & 4.47  & 70.22 & 8.47 &  71.23 & 7.54  \\
\rowcolor{wacvblue!15}
Ours (ImageBind)               & \textbf{91.48} & \textbf{0.52} & \textbf{84.94} & \textbf{0.06} & \textbf{86.29} & \textbf{0.44} \\

\midrule
\bottomrule
\end{tabular}
}
\end{table}

%% file: Tables/ablation_loss_upd.tex
\begin{table}[ht]
\centering
    \caption{Effectiveness of different loss components ($\mathcal{L}_{ce}$, $\mathcal{L}_{fd}$, and $\mathcal{L}_{kd}$) on AVE-CI dataset. \textbf{Bold} denotes the best results.}
\label{tab:ablation_loss_upd}
\resizebox{.8\columnwidth}{!}{
\centering
\setlength{\tabcolsep}{5pt}
\begin{tabular}{c c c c c c c c c}
\toprule \midrule

\multirow{2}{*}{$\mathcal{L}_{ce}$} & \multirow{2}{*}{$\mathcal{L}_{kd}$} & \multirow{2}{*}{$\mathcal{L}_{fd}$} & \multicolumn{2}{c}{SAM-Audio} & \multicolumn{2}{c}{ImageBind} & \multicolumn{2}{c}{Qwen2.5} \\
\cmidrule(lr){4-5} \cmidrule(lr){6-7} \cmidrule(lr){8-9}
& & & $\mathcal{A}_m \uparrow$ & $\mathcal{F}_{avg} \downarrow$& $\mathcal{A}_m \uparrow$& $\mathcal{F}_{avg} \downarrow$& $\mathcal{A}_m \uparrow$& $\mathcal{F}_{avg} \downarrow$ \\
\midrule
\cmark & \xmark  & \xmark & 87.26 & 4.29 & 86.55 & 3.18 & 83.20 & 7.32 \\
\cmark & \cmark & \xmark & 87.34 & 2.30 & 87.60 & 2.80 & 83.94 & 5.45 \\
\cmark & \xmark     & \cmark  & 87.79 & 1.32 & 87.78 & 1.02 & 84.49 & 5.01 \\ 
\rowcolor{wacvblue!10}
\cmark &   \cmark &  \cmark  & \textbf{88.68} & \textbf{0.42} & \textbf{91.48} & \textbf{0.52} & \textbf{86.14} & \textbf{4.18} \\
\midrule \bottomrule
\end{tabular}
}
\end{table}

%% file: Tables/ablation_hyp.tex
\begin{table}[htbp]
\centering
\caption{Sensitivity analysis of distillation hyperparameters $\lambda_{fd}$ and $\lambda_{kd}$ by comparing the accuracy ($\mathcal{A}_m$) for different MFM-based backbones. \textbf{Bold} denotes the adopted configuration.}
\label{tab:ablation_hyp}
\resizebox{.8\columnwidth}{!}{
\setlength{\tabcolsep}{1pt}
\begin{tabular}{c c c c c c c c}
\toprule \midrule
\multirow{2}{*}{$\lambda_{fd}$} & \multirow{2}{*}{$\lambda_{kd}$} & \multicolumn{3}{c}{AVE-CI}  & \multicolumn{3}{c}{K-S-CI} \\
\cmidrule(lr){3-5} \cmidrule(lr){6-8} 
&  & SAM-Audio & ImageBind & Qwen2.5 & SAM-Audio & ImageBind & Qwen2.5\\
\midrule
0.2  &  1.0     & 88.04 & 88.66 & 82.92 & \textbf{84.50} & 84.88 & 81.52 \\ 
0.4  &  1.0     & 87.90 & 89.33 & 82.68 & 84.36 & 85.13 & 81.52 \\
0.6  &  1.0     & 88.06 & 89.48 & 82.41 & 84.07 & 85.11 & 81.49 \\
0.8  &  1.0     & 87.92 & 89.36 & 82.57 & 83.72 & 85.05 & 81.46 \\
\rowcolor{wacvblue!10}
\textbf{1.0}  &  \textbf{1.0}     & \textbf{88.68} & \textbf{91.48} & \textbf{86.14} & 84.01 & \textbf{86.29} & \textbf{81.85} \\
0.5  &  0.8     & 88.14 & 89.40 & 82.81 & 84.35 & 85.10 & 81.53 \\ 
0.5  &  0.6     & 88.06 & 89.51 & 82.83 & 84.28 & 85.47 & 80.57  \\
0.5  &  0.4     & 88.09 & 89.39 & 82.80 & 84.33 & 85.67 & 80.64 \\
0.5  &  0.2     & 88.44 & 89.96 & 82.68 & 84.04 & 85.61 & 81.71 \\
\midrule \bottomrule
\end{tabular}
}
\end{table}

%% file: sections/5_Conclusion.tex
\section{Conclusion}

In this work, we addressed the challenge of Class-Incremental Learning (CIL) by leveraging Multimodal Foundational Models (MFMs). 
By leveraging the frozen priors of MFMs such as \textit{SAM-Audio}, \textit{ImageBind}, and \textit{Qwen2.5-Omni}, we propose a unified, shared latent strategy and present a generic CIL system for audio-visual class recognition. Furthermore, we employ an Attention-based Contextual Guidance module that uses audio representations as context, followed by dual distillation, for effective class recognition. Empirical results across multiple benchmarks show that our method significantly outperforms current baselines. 

%% file: sections/5.5_Acknowledgement.tex
\section{Acknowledgement}

This work was started during the internship of Avi Gupta at Dolby Laboratories and was later supported in part by the Infosys Centre for Artificial Intelligence, IIIT-Delhi, and in part by the Institute Fellowship of the IIIT-Delhi. A preliminary version of this work was presented at the ICML workshop~\cite{DBLP:journals/corr/abs-2606-10887}.

%% file: sections/6_Supplementary.tex
\appendix
\section{Introduction}
In this section, we present the additional analysis on our proposed ``Generalizing Multimodal Foundational Models in Context-Driven Class Incremental Learning''. We first present the training pipeline of our proposed work (Algorithm\textbf{~\ref{algo:main}}), followed by ablation studies on individual modalities, analysis across multiple runs, and task-wise performance across different settings.

\input{Algorithm/pseudo}

\section{\textbf{Effectiveness of Individual Modality}}

\input{Tables/ablation_modality}To assess the necessity of a multimodal framework for class-incremental learning, we conduct an ablation study by isolating the contributions of audio and visual inputs. In Table\textbf{~\ref{tab:ablation_2}}, unimodal baselines demonstrate a significant performance deficit, failing to capture the complex semantic nuances. Furthermore, we observe a performance improvement through the integration of both modalities. For instance, in the \textit{SAM-Audio} backbone, fusing audio and video yields a significant improvement in Mean Accuracy over the best-performing unimodal variant. 
As for \textit{Qwen2.5}, we observe a high unimodal performance for both audio \textbf{(67.75\%)} and visual-only \textbf{(69.84\%)} counterparts. Nevertheless, across these varying priors, the dual-modality configuration consistently emerges as optimal, achieving a peak accuracy of \textbf{86.14\%}.

\input{Tables/multiple_runs}
\section{Analysis over Multiple Runs}
For robustness across different random runs, we have run three independent random seeds on AVE-CI, VS100-CI, and K-S-CI for all three MFMs and the conventional AudioMAE/VideoMAE backbone, and report mean and standard deviation in Table~\ref{tab:multiple_runs}. The main paper reports the single run, and the supplementary presents the seed-averaged values. The standard deviations are small, and the Ours--AV-CIL margin exceeds the combined variance at every backbone, indicating the improvement is robust to task ordering rather than an artifact of a particular seed.

\section{Task-wise Performance for different settings}
We present the task-wise performance (Table.\textbf{~\ref{tab:Supp_1}-~\ref{tab:supp_6}}) across diverse task sequences (briefly described in $\S$4.2 of the paper). We evaluate the corresponding performance across the three foundational models and compare against AV-CIL. Our proposed method consistently outperforms the AV-CIL baseline across all three datasets in both traditional and task-wise learning settings. Among the backbones, ImageBind establishes the highest performance; in the traditional Kinetics-CI setting, it achieves 96.77\% on the initial task and sustains 81.05\% by the final step. On the AVE-CI dataset, our method preserves near-perfect accuracy on the initial task across all backbones. Even in the highly challenging 20-task split of VGGSound100-CI, our approach prevents the severe performance collapse observed in the baseline, keeping late-stage task accuracies firmly in the 65\% to 75\% range.

\input{Tables/mem_analysis}
\section{Compute/Memory}
Our framework extracts audio/visual features from the frozen encoders of MFMs, which are computed offline; later, incremental training and inference are performed on these features via a contextual guidance module and a linear classifier. Table\textbf{~\ref{tab:mem_analysis}} reports memory overhead for each backbone. 
The trainable parameters for MFM-based backbones are \textbf{$\approx$70M}, their FLOPs are on the order of \textbf{$\approx$1.0GFLOPs}, and peak training VRAM is \textbf{$\approx$3.5GB} is determined by the shallow trainable module rather than the encoders. 
The exemplar buffer stores features rather than raw media, significantly reducing per-buffer storage.
The exemplar cost (averaged across different MFM backbones) for AVE-CI (340 samples) is \textbf{255MB}, VS100-CI (1500 samples) is \textbf{1.13GB}, and K-S-CI (500 samples) is \textbf{375MB}. 
This design makes the recurring incremental cost lightweight, a practical advantage for continual deployment.
\input{Tables/Supp_2}
\input{Tables/Supp_6}

%% file: Algorithm/pseudo.tex
\begin{algorithm}[htbp]
\caption{Training Pipeline}
\label{algo:main}
\begin{algorithmic}[1]
\State \textbf{Input:} Training dataset for current task $t$, $\mathcal{D}'_t=\mathcal{D}_t\cup\mathcal{M}_{t-1}$; total tasks $T$.
\State \textcolor{gray}{// $model\in\{sam, ibind, qwen\}$}
\For{$t \in \{0,1,\dots,T\}$}
\For{$x_{a,i},\ x_{v,i},\ y_i \in \mathcal{D}'_t$}
\State \textcolor{gray}{// -- Foundational Multimodal Feature Extraction ---}

\State $z^{model}_{a,i}, z^{model}_{v,i}\leftarrow\ $Features($x_a, x_v; model$) 
\State \textcolor{gray}{// --- Attention-based Contextual Guidance ---}
\State $z^{'model}_{v,i} \leftarrow\ $Guided Attention($z^{model}_{v,i},\ z^{model}_{a,i}$) 
\State \textcolor{gray}{// --- Classifier ---}
\State $\hat{y}_i \leftarrow\ $Classifier($z^{'model}_{v,i}$)
\If{$t=0$}
\State Loss $\leftarrow\ \mathcal{L}_{ce}$ 
\Else
\State Loss $\leftarrow\ \mathcal{L}_{tot}$ 
\EndIf
\State Loss($\hat{y}_i, y_i$)
\EndFor
\EndFor
\end{algorithmic}
\end{algorithm}

%% file: Tables/ablation_modality.tex
\begin{table*}[ht]
\centering
\caption{Effectiveness of individual modality on the AVE dataset. \textbf{Bold} denotes the best accuracy ($\mathcal{A}_m$).}
\label{tab:ablation_2}
\resizebox{.4\textwidth}{!}{
\centering
\setlength{\tabcolsep}{7pt}
\begin{tabular}{c c c c}
\toprule
\midrule
Audio & Video & SAM-Audio & Qwen2.5 \\
\midrule
\cmark &\xmark  & 30.75 & 67.75   \\ 
\xmark & \cmark & 38.98 & 69.84    \\
\rowcolor{wacvblue!10}
\cmark & \cmark & \textbf{88.68} & \textbf{86.14} \\
\midrule
\bottomrule
\end{tabular}
}
\end{table*}

%% file: Tables/multiple_runs.tex
\begin{table*}[ht]
\centering
\caption{Analysis over three random seeds for three benchmark datasets.}
\label{tab:multiple_runs}
\renewcommand{\arraystretch}{1.1}
\resizebox{\textwidth}{!}{
\setlength{\tabcolsep}{5pt}
\begin{tabular}{l c c >{\columncolor{wacvblue!10}} c >{\columncolor{wacvblue!10}} c c c >{\columncolor{wacvblue!10}} c >{\columncolor{wacvblue!10}} c c c >{\columncolor{wacvblue!10}} c >{\columncolor{wacvblue!10}} c}
\toprule
\midrule
\multirow{3}{*}{Methods} & \multicolumn{4}{c}{AVE-CI} & \multicolumn{4}{c}{VS100-CI} & \multicolumn{4}{c}{K-S-CI} \\
\cmidrule(lr){2-5} \cmidrule(lr){6-9} \cmidrule(lr){10-13}

 & \multicolumn{2}{c}{AV-CIL} & \multicolumn{2}{>{\columncolor{wacvblue!10}}c}{Ours} & \multicolumn{2}{c}{AV-CIL} & \multicolumn{2}{>{\columncolor{wacvblue!10}}c}{Ours} & \multicolumn{2}{c}{AV-CIL} & \multicolumn{2}{>{\columncolor{wacvblue!10}}c}{Ours} \\
 
\cmidrule(lr){2-3} \cmidrule(lr){4-5} \cmidrule(lr){6-7} \cmidrule(lr){8-9} \cmidrule(lr){10-11} \cmidrule(lr){12-13}

& $\mathcal{A}_m$ & $\mathcal{F}_{avg}$ & $\mathcal{A}_m$ & $\mathcal{F}_{avg}$ & $\mathcal{A}_m$ & $\mathcal{F}_{avg}$ & $\mathcal{A}_m$ & $\mathcal{F}_{avg}$& $\mathcal{A}_m$ & $\mathcal{F}_{avg}$ & $\mathcal{A}_m$ & $\mathcal{F}_{avg}$ \\ 
\midrule
AudioMAE/VideoMAE &74.04 & 7.63 & 78.06$\pm$1.17&1.12$\pm$0.45 &72.80 &5.49&79.88$\pm$1.37&4.65$\pm$0.55&73.06 &6.48 &80.24$\pm$1.32&5.45$\pm$0.76 \\
SAM-Audio &84.38 & 2.24&88.72$\pm$0.67&0.43$\pm$0.03  &74.04 &4.11&79.11$\pm$0.76&0.37$\pm$0.08&80.21 &1.45&83.34$\pm$1.75&0.54$\pm$0.12\\
ImageBind &85.80 &3.04&91.25$\pm$0.51&0.55$\pm$0.06&77.70 &7.72&84.01$\pm$1.21&0.10$\pm$0.06& 77.96  &7.09&86.11$\pm$0.55&0.76$\pm$0.54\\
Qwen2.5 &82.98 &5.91&86.97$\pm$1.12&4.66$\pm$0.97&80.43 &5.40&82.50$\pm$0.24&1.34$\pm$0.32&79.32 &4.28&82.20$\pm$0.44&3.54$\pm$0.40\\
\midrule
\bottomrule
\end{tabular}
}
\end{table*}

%% file: Tables/mem_analysis.tex
\begin{table*}[ht]
\centering
\caption{Compute/Memory overhead for different backbones.}
\label{tab:mem_analysis}
\renewcommand{\arraystretch}{1.1}
\resizebox{.65\textwidth}{!}{
\setlength{\tabcolsep}{5pt}
\begin{tabular}{l c c c c c c}
\toprule
\midrule
Backbone & Frozen Params & Trainable Params & GFLOPs & Latency & VRAM \\ \midrule

SAMAudio  &  $\approx$500M & 27.31M & 1.06G & 4.11ms & 1.58GB \\
ImageBind&  $\approx$700M & 27.31M & 0.54G & 5.84ms & 1.41GB\\
Qwen2.5& $\approx$1.5B  & 109.15M & 2.17G & 6.12ms & 5.61GB\\
\midrule
\bottomrule
\end{tabular}
}
\end{table*}

%% file: Tables/Supp_2.tex
\begin{table*}[ht]
\centering
\caption{Comparison of our proposed approach on different MFMs with AV-CIL on AVE-CI dataset \textit{\textbf{4 Tasks (7 classes per task)}}. Best results ($\%$) are marked in \textbf{Bold}.}
\label{tab:Supp_1}
\renewcommand{\arraystretch}{1}
\resizebox{0.65\textwidth}{!}{
\setlength{\tabcolsep}{8pt}
\begin{tabular}{lcccccc}
\toprule \midrule
\multirow{2}{*}{Steps} & \multicolumn{2}{c}{\textit{\textbf{SAMAudio}}} & \multicolumn{2}{c}{\textit{\textbf{ImageBind}}} & \multicolumn{2}{c}{\textit{\textbf{Qwen2.5}}} \\
\cmidrule(lr){2-3} \cmidrule(lr){4-5} \cmidrule(lr){6-7}
& AV-CIL & \textbf{Ours} & AV-CIL & \textbf{Ours} & AV-CIL & \textbf{Ours} \\
\midrule
Step 0 & 86.15 & \textbf{90.39} & 91.34 & \textbf{95.20} & 90.38 & \textbf{92.30} \\
Step 1 & 85.23 & \textbf{91.43} & 87.14 & \textbf{95.71} & 83.33 & \textbf{86.19} \\
Step 2 & 83.01 & \textbf{86.35} & 84.75 & \textbf{86.67} & 81.58 & \textbf{83.80} \\
Step 3 & 83.14 & \textbf{86.55} & 79.90 & \textbf{88.32} & 76.64 & \textbf{82.23} \\ \midrule
\bottomrule
\end{tabular}
}
\end{table*}
\begin{table*}[ht]
\centering
\caption{Comparison of our proposed approach on different MFMs with AV-CIL on the VGGSound100-CI dataset \textit{\textbf{10 Tasks (10 classes per task)}}. Best results ($\%$) are marked in \textbf{Bold}.}
\label{tab:Supp_2}
\renewcommand{\arraystretch}{1}
\resizebox{0.65\textwidth}{!}{
\setlength{\tabcolsep}{10pt}
\begin{tabular}{l c c c c c c}
\toprule \midrule
\multirow{2}{*}{Steps} & \multicolumn{2}{c}{\textbf{\textit{SAMAudio}}} & \multicolumn{2}{c}{\textbf{\textit{ImageBind}}} & \multicolumn{2}{c}{\textbf{\textit{Qwen2.5}}} \\
\cmidrule(lr){2-3} \cmidrule(lr){4-5} \cmidrule(lr){6-7}
& AV-CIL & \textbf{Ours} & AV-CIL & \textbf{Ours} & AV-CIL & \textbf{Ours} \\
\midrule
Step 0 & 83.14 & \textbf{91.33} & 92.33 & \textbf{93.34} & 82.54 & \textbf{92.94} \\
Step 1 & 81.84 & \textbf{85.70} & 86.30 & \textbf{89.02} & 80.93 & \textbf{86.20} \\
Step 2 & 78.06 & \textbf{83.03} & 84.10 & \textbf{87.46} & 80.47 & \textbf{87.19} \\
Step 3 & 76.45 & \textbf{82.02} & 81.12 & \textbf{86.91} & 77.0 & \textbf{83.59} \\
Step 4 & 76.36 & \textbf{80.61} & 78.80 & \textbf{86.24} & 75.31 & \textbf{83.35} \\
Step 5 & 72.69 & \textbf{77.36} & 75.62 & \textbf{82.96} & 71.35 & \textbf{81.25} \\
Step 6 & 70.05 & \textbf{75.65} & 73.18 & \textbf{82.43} & 70.31 & \textbf{79.88} \\
Step 7 & 68.93 & \textbf{75.21} & 71.33 & \textbf{81.96} & 67.25 & \textbf{78.51} \\
Step 8 & 67.15 & \textbf{72.82} & 67.27 & \textbf{79.64} & 65.01 & \textbf{75.93} \\
Step 9 & 65.76 & \textbf{72.60} & 66.86 & \textbf{79.46} & 64.14 & \textbf{75.20} \\ \midrule
\bottomrule
\end{tabular}
}
\end{table*}
\begin{table*}[ht]
\centering
\caption{Comparison of our proposed approach on different MFMs with AV-CIL on Kinetics-CI dataset \textit{\textbf{5 Tasks (6 classes per task)}}. Best results ($\%$) are marked in \textbf{Bold}.}
\label{tab:Supp_3}
\renewcommand{\arraystretch}{1}
\resizebox{0.65\textwidth}{!}{
\setlength{\tabcolsep}{13pt}
\begin{tabular}{l c c c c c c}
\toprule \midrule
\multirow{2}{*}{Steps} & \multicolumn{2}{c}{\textbf{\textit{SAMAudio}}} & \multicolumn{2}{c}{\textbf{\textit{ImageBind}}} & \multicolumn{2}{c}{\textbf{\textit{Qwen2.5}}} \\
\cmidrule(lr){2-3} \cmidrule(lr){4-5} \cmidrule(lr){6-7}
& AV-CIL & \textbf{Ours} & AV-CIL & \textbf{Ours} & AV-CIL & \textbf{Ours} \\
\midrule
Step 0 & 89.59 & \textbf{95.97} & 93.14 & \textbf{96.77} & 92.74 & \textbf{94.09} \\
Step 1 & 80.05 & \textbf{85.97} & 77.97 & \textbf{85.71} & 77.77 & \textbf{81.95} \\
Step 2 & 80.23 & \textbf{80.85} & 76.78 & \textbf{84.75} & 77.57 & \textbf{80.24} \\
Step 3 & 79.27 & \textbf{79.42} & 73.98 & \textbf{83.17} & 76.46 & \textbf{78.64} \\
Step 4 & 74.23 & \textbf{77.83} & 67.93 & \textbf{81.05} & 72.04 & \textbf{74.31} \\ \midrule
\bottomrule
\end{tabular}
}
\end{table*}

%% file: Tables/Supp_6.tex
\begin{table*}[ht]
\centering
\caption{Task-wise Comparison of our proposed approach on different MFMs with AV-CIL on AVE-CI dataset. Best results ($\%$) are marked in \textbf{Bold}.}
\label{tab:supp_3}
\renewcommand{\arraystretch}{1}
\resizebox{0.65\textwidth}{!}{
\setlength{\tabcolsep}{5pt}
\begin{tabular}{l c c c c c c}
\toprule \midrule
\multirow{2}{*}{Steps} & \multicolumn{2}{c}{\textbf{\textit{SAMAudio}}} & \multicolumn{2}{c}{\textbf{\textit{ImageBind}}} & \multicolumn{2}{c}{\textbf{\textit{Qwen2.5}}} \\
\cmidrule(lr){2-3} \cmidrule(lr){4-5} \cmidrule(lr){6-7}
& AV-CIL & \textbf{Ours} & AV-CIL & \textbf{Ours} & AV-CIL & \textbf{Ours} \\
\midrule
\textbf{\textit{7 Tasks (4 classes per task)}}\\
\cmidrule(lr){1-1}
Step 0 & 86.55 & \textbf{99.2} & 90.00 & \textbf{98.89} & 88.00 & \textbf{98.28} \\
Step 1 & 86.36 & \textbf{70.0} & 77.00 & \textbf{80.0} & 55.45 & \textbf{63.63} \\
Step 2 & 85.45 & \textbf{86.93} & 75.34 & \textbf{91.47} & 78.07 & \textbf{78.98} \\
Step 3 & 83.83 & \textbf{88.07} & 74.36 & \textbf{92.59} & 62.30 & \textbf{68.72} \\
Step 4 & 81.15 & \textbf{82.62} & 74.59 & \textbf{86.88} & 68.69 & \textbf{71.48} \\
Step 5 & 85.18 & \textbf{83.57} & 73.00 & \textbf{86.40} & 69.89 & \textbf{70.82} \\
Step 6 & 83.91 & \textbf{82.49} & 71.98 & \textbf{85.53} & 65.38 & \textbf{73.10} \\
\midrule
\textit{\textbf{2 Tasks (14 classes per task)}}\\
\cmidrule(lr){1-1}
Step 0 & 85.71 & \textbf{94.76} & 82.38 & \textbf{93.80} & 78.10 & \textbf{80.05} \\
Step 1 & 79.34 & \textbf{85.03} & 73.50 & \textbf{89.85} & 69.44 & \textbf{79.95} \\ \midrule
\bottomrule
\end{tabular}
}
\end{table*}
\begin{table*}[ht]
\centering
\caption{Task-wise Comparison of our proposed approach on different MFMs with AV-CIL on VGGSound100-CI dataset. Best results ($\%$) are marked in \textbf{Bold}.}
\label{tab:supp_5}
\renewcommand{\arraystretch}{1}
\resizebox{0.65\textwidth}{!}{
\setlength{\tabcolsep}{4pt}
\begin{tabular}{l c c c c c c}
\toprule \midrule
\multirow{2}{*}{Steps} & \multicolumn{2}{c}{\textbf{\textit{SAMAudio}}} & \multicolumn{2}{c}{\textbf{\textit{ImageBind}}} & \multicolumn{2}{c}{\textbf{\textit{Qwen2.5}}} \\
\cmidrule(lr){2-3} \cmidrule(lr){4-5} \cmidrule(lr){6-7}
& AV-CIL & \textbf{Ours} & AV-CIL & \textbf{Ours} & AV-CIL & \textbf{Ours} \\
\midrule
\textit{\textbf{5 Tasks (20 classes per task)}}\\
\cmidrule(lr){1-1}
Step 0 & 84.36 & \textbf{91.74} & 91.74 & \textbf{93.66} & 82.35 & \textbf{88.78} \\
Step 1 & 79.38 & \textbf{80.51} & 81.04 & \textbf{83.38} & 76.25 & \textbf{72.92} \\
Step 2 & 73.20 & \textbf{75.65} & 76.93 & \textbf{79.28} & 71.62 & \textbf{85.89} \\
Step 3 & 70.17 & \textbf{75.16} & 71.44 & \textbf{77.76} & 67.83 & \textbf{82.46} \\
Step 4 & 66.87 & \textbf{73.54} & 67.02 & \textbf{75.98} & 64.35 & \textbf{90.04} \\
\midrule
\textit{\textbf{2 Tasks (50 classes per task)}}\\
\cmidrule(lr){1-1}
Step 0 & 79.94 & \textbf{87.33} & 87.13 & \textbf{91.19} & 78.33 & \textbf{82.48} \\
Step 1 & 68.70 & \textbf{72.74} & 70.32 & \textbf{77.55} & 63.93 & \textbf{81.80} \\
\midrule
\textit{\textbf{20 Tasks (5 classes per task)}}\\
\cmidrule(lr){1-1}
Step 0 & 78.71 & \textbf{85.48} & 77.90 & \textbf{87.90} & 80.32 & \textbf{82.56} \\
Step 1 & 83.55 & \textbf{88.50} & 87.50 & \textbf{89.52} & 81.33 & \textbf{83.59} \\
Step 2 & 83.94 & \textbf{88.02} & 88.29 & \textbf{90.71} & 73.27 & \textbf{77.74} \\
Step 3 & 81.04 & \textbf{83.08} & 83.59 & \textbf{87.92} & 70.94 & \textbf{80.50} \\
Step 4 & 80.34 & \textbf{81.16} & 83.01 & \textbf{86.80} & 70.42 & \textbf{79.49} \\
Step 5 & 77.86 & \textbf{78.20} & 81.76 & \textbf{83.84} & 71.48 & \textbf{78.05} \\
Step 6 & 79.69 & \textbf{79.45} & 82.04 & \textbf{84.97} & 69.75 & \textbf{77.46} \\
Step 7 & 76.35 & \textbf{77.20} & 79.15 & \textbf{82.53} & 64.94 & \textbf{76.18} \\
Step 8 & 77.25 & \textbf{77.20} & 79.07 & \textbf{83.72} & 66.85 & \textbf{76.03} \\
Step 9 & 76.08 & \textbf{75.70} & 78.88 & \textbf{82.98} & 64.67 & \textbf{75.03} \\
Step 10 & 73.64 & \textbf{73.17} & 76.29 & \textbf{79.62} & 63.02 & \textbf{74.51} \\
Step 11 & 71.59 & \textbf{72.23} & 74.71 & \textbf{78.71} & 61.19 & \textbf{71.93} \\
Step 12 & 69.75 & \textbf{69.78} & 73.55 & \textbf{76.71} & 69.96 & \textbf{71.41} \\
Step 13 & 67.84 & \textbf{70.10} & 71.11 & \textbf{76.52} & 69.33 & \textbf{71.07} \\
Step 14 & 67.96 & \textbf{68.16} & 70.47 & \textbf{75.87} & 66.83 & \textbf{69.86} \\
Step 15 & 58.81 & \textbf{67.74} & 69.43 & \textbf{74.81} & 65.57 & \textbf{92.95} \\
Step 16 & 57.35 & \textbf{65.06} & 67.71 & \textbf{72.40} & 64.03 & \textbf{78.25} \\
Step 17 & 56.35 & \textbf{65.10} & 65.86 & \textbf{72.60} & 64.30 & \textbf{79.68} \\
Step 18 & 55.54 & \textbf{63.78} & 64.77 & \textbf{70.90} & 63.46 & \textbf{78.23} \\
Step 19 & 54.81 & \textbf{64.06} & 63.78 & \textbf{71.93} & 62.40 & \textbf{75.28} \\ \midrule
\bottomrule
\end{tabular}
}
\end{table*}
\begin{table*}[ht]
\centering
\caption{Task-wise Comparison of our proposed approach on different MFMs with AV-CIL on Kinetics-CI dataset. Best results ($\%$) are marked in \textbf{Bold}.}
\label{tab:supp_6}
\renewcommand{\arraystretch}{1}
\resizebox{0.65\textwidth}{!}{
\setlength{\tabcolsep}{4pt}
\begin{tabular}{l c c c c c c}
\toprule
\midrule
\multirow{2}{*}{Steps} & \multicolumn{2}{c}{\textbf{\textit{SAMAudio}}} & \multicolumn{2}{c}{\textbf{\textit{ImageBind}}} & \multicolumn{2}{c}{\textbf{\textit{Qwen2.5}}} \\
\cmidrule(lr){2-3} \cmidrule(lr){4-5} \cmidrule(lr){6-7}
& AV-CIL & \textbf{Ours} & AV-CIL & \textbf{Ours} & AV-CIL & \textbf{Ours} \\
\midrule
\textit{\textbf{6 Tasks (5 classes per task)}}\\
\cmidrule(lr){1-1}
Step 0 & 89.19 & \textbf{95.82} & 92.74 & \textbf{98.12} & 84.35 & \textbf{95.18} \\
Step 1 & 81.42 & \textbf{86.09} & 77.58 & \textbf{86.75} & 70.90 & \textbf{83.28} \\
Step 2 & 70.77 & \textbf{76.04} & 76.39 & \textbf{85.62} & 69.68 & \textbf{76.04} \\
Step 3 & 79.37 & \textbf{81.75} & 72.00 & \textbf{82.91} & 66.76 & \textbf{79.35} \\
Step 4 & 74.86 & \textbf{79.23} & 66.15 & \textbf{80.85} & 61.66 & \textbf{77.31} \\
Step 5 & 72.66 & \textbf{78.14} & 63.80 & \textbf{79.21} & 59.66 & \textbf{73.75} \\
\midrule
\textit{\textbf{2 Tasks (15 classes per task)}}\\
\cmidrule(lr){1-1}
Step 0 & 84.04 & \textbf{87.34} & 85.11 & \textbf{91.70} & 71.82 & \textbf{80.54} \\
Step 1 & 77.89 & \textbf{81.05} & 76.48 & \textbf{83.81} & 62.67 & \textbf{76.10} \\
\midrule
\textit{\textbf{15 Tasks (2 classes per task)}}\\
\cmidrule(lr){1-1}
Step 0 & 90.00 & \textbf{99.15} & 97.67 & \textbf{99.78} & 88.84 & \textbf{99.15} \\
Step 1 & 87.59 & \textbf{94.26} & 86.75 & \textbf{96.72} & 68.31 & \textbf{88.11} \\
Step 2 & 85.97 & \textbf{87.63} & 73.39 & \textbf{94.35} & 74.27 & \textbf{79.03} \\
Step 3 & 85.09 & \textbf{86.60} & 75.46 & \textbf{88.60} & 74.36 & \textbf{80.40} \\
Step 4 & 78.92 & \textbf{88.28} & 72.41 & \textbf{88.28} & 72.78 & \textbf{81.09} \\
Step 5 & 79.28 & \textbf{86.88} & 67.84 & \textbf{87.40} & 67.00 & \textbf{78.96} \\
Step 6 & 69.61 & \textbf{79.44} & 66.83 & \textbf{82.67} & 68.56 & \textbf{76.33} \\
Step 7 & 67.19 & \textbf{76.60} & 63.92 & \textbf{83.69} & 62.61 & \textbf{75.24} \\
Step 8 & 72.88 & \textbf{80.68} & 60.82 & \textbf{81.63} & 63.35 & \textbf{75.22} \\
Step 9 & 67.90 & \textbf{77.48} & 59.19 & \textbf{81.37} & 63.90 & \textbf{74.53} \\
Step 10 & 69.78 & \textbf{78.90} & 61.61 & \textbf{81.50} & 65.70 & \textbf{74.75} \\
Step 11 & 66.59 & \textbf{77.09} & 58.99 & \textbf{77.80} & 61.90 & \textbf{71.20} \\
Step 12 & 68.40 & \textbf{77.92} & 60.83 & \textbf{80.18} & 61.80 & \textbf{71.63} \\
Step 13 & 67.93 & \textbf{77.52} & 56.41 & \textbf{79.00} & 63.51 & \textbf{71.03} \\
Step 14 & 64.39 & \textbf{74.92} & 57.07 & \textbf{76.97} & 59.95 & \textbf{67.82} \\
\midrule
\bottomrule
\end{tabular}
}
\end{table*}